\documentclass[runningheads]{llncs}
\usepackage[T1]{fontenc}
\usepackage{graphicx,verbatim}
\usepackage{multirow}
\usepackage{amsmath}
\usepackage{amssymb}
\usepackage{array}
\usepackage{graphicx}
\usepackage{hyperref}
\begin{document}

\title{Trexplorer Super: Topologically Correct Centerline Tree Tracking of Tubular Objects in CT Volumes}
\titlerunning{Trexplorer Super}
\author{Roman Naeem
\and David Hagerman
\and Jennifer Alvén
\and Lennart Svensson
\and Fredrik Kahl}
\institute{Chalmers University of Technology, 412 96 Gothenburg, Sweden\\
\email{nroman@chalmers.se}}

\maketitle

\begin{abstract}
Tubular tree structures, such as blood vessels and airways, are essential in human anatomy and accurately tracking them while preserving their topology is crucial for various downstream tasks. Trexplorer is a recurrent model designed for centerline tracking in 3D medical images but it struggles with predicting duplicate branches and terminating tracking prematurely. To address these issues, we present Trexplorer Super, an enhanced version that notably improves performance through novel advancements. However, evaluating centerline tracking models is challenging due to the lack of public datasets. To enable thorough evaluation, we develop three centerline datasets, one synthetic and two real, each with increasing difficulty. Using these datasets, we conduct a comprehensive evaluation of existing state-of-the-art (SOTA) models and compare them with our approach. Trexplorer Super outperforms previous SOTA models on every dataset. Our results also highlight that strong performance on synthetic data does not necessarily translate to real datasets. The code and datasets are available at \\ \href{https://github.com/RomStriker/Trexplorer-Super}
{https://github.com/RomStriker/Trexplorer-Super}.

\keywords{centerline tracking  \and tubular structures \and tree topology.}

\end{abstract}

\section{Introduction}

Tubular tree structures in the vascular and respiratory systems play a critical role for transporting essential substances throughout the body. Accurately tracking the centerlines of these structures in medical images is fundamental for early diagnosis, treatment, and various downstream tasks~\cite{miraucourt2017blood,huang2011interactive,khan2018three,choi2021ct}. In this paper, we introduce a new method for centerline tree tracking and propose a comprehensive framework for its evaluation.

Several existing approaches address the challenge of centerline extraction, but they each come with limitations. A common approach segments the image and then applies skeletonization~\cite{tetteh2020deepvesselnet}, but these models struggle with long-range dependencies, leading to connectivity issues. Other models~\cite{prabhakar2024vesselformer,shit2022relationformer} detect centerline nodes and edges in a two-step process but also suffer from connectivity errors. Recurrent models, such as reinforcement learning-based methods~\cite{zhang2020branch,li2021deep}, iteratively track centerlines but rely on complex pipelines. Trexplorer~\cite{naeem2024trexplorer} simplifies this with a DETR-based transformer~\cite{carion2020end} that uses breadth-first tracking which ensures correct topology. However, it struggles with duplicate branch detections and premature tracking terminations.

To overcome the limitations of existing centerline tracking methods, we propose Trexplorer Super, which builds on the Trexplorer model with several key enhancements to improve accuracy, robustness, and completeness. Our method reduces premature terminations and duplicate branches while improving new branch detection and preserving fine spatial details in image features. To ensure more consistent centerline extraction, we introduce Super Trajectory Training, a strategy that retains and reuses tracking information across multiple steps. Additionally, we refine feature representation with Focal Cross Attention, which selectively attends to a focal region in high-resolution image features while maintaining broader contextual awareness. To further enhance robustness, we employ Target Augmentation, a strategy that improves bifurcation and new branch detection while minimizing duplicate branches. These advancements contribute to a more reliable and comprehensive centerline tracking framework.

Evaluating centerline tracking in 3D medical images is challenging due to the lack of publicly available real datasets. Existing synthetic datasets have topological limitations, and strong performance on them does not generalize well to real data. To address this, we create one synthetic and two real datasets and establish a comprehensive baseline by evaluating prior SOTA models and our approach using point, branch, and tree-level metrics.

Our key contributions are: \textbf{(1)} Enhancing the Trexplorer framework with novel techniques: Super Trajectory Training, Focal Cross Attention, and Target Augmentation. \textbf{(2)} Creating three datasets and thoroughly evaluating the previous SOTA models and our method.

\section{Method}

Our goal is to estimate the centerline tree from a given CT volume and an initial starting point. The centerline tree is represented as a graph $(V,E)$ with $V$ nodes and $E$ edges. Each node ${\bf v} \in V$ is defined as a vector ${\bf v}=[x,y,z,r]$ representing the 3D position and radius of a centerline point, while an edge ${\bf e} \in E$ is a connection between two nodes. 

\subsubsection{Trexplorer Super Architecture.}
\begin{figure}[t!] 
    \centering
    \includegraphics[width=1.00\textwidth]{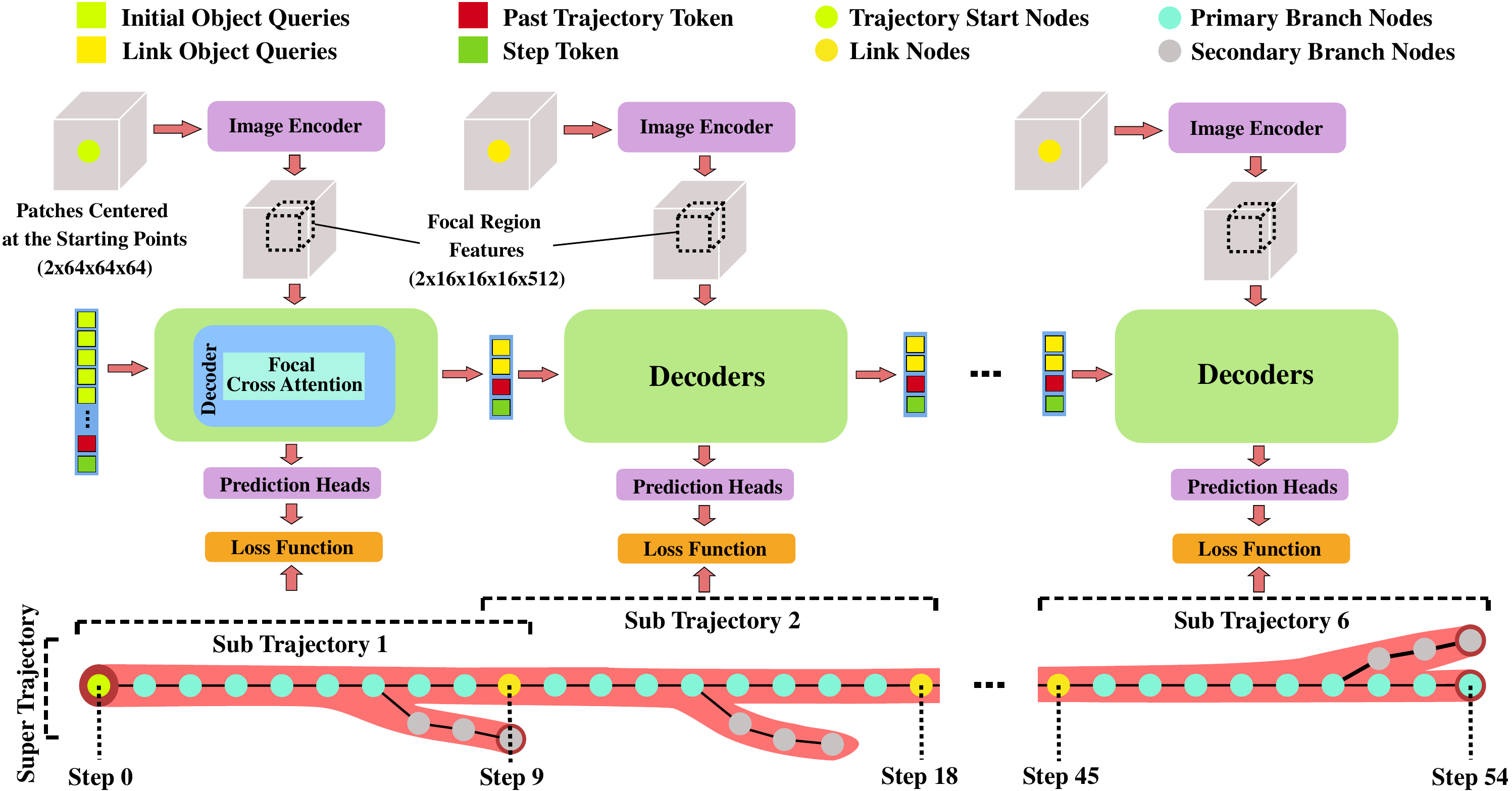}
    \caption{Super Trajectory Training for Trexplorer Super using Focal Cross Attention.}
    \label{fig:arch}
\end{figure}

Trexplorer Super is a DETR-based model and uses object queries to track branches. It begins at the root node and tracks each branch until it reaches a leaf (end) node. It estimates the total number of branches and the number of nodes within each branch. The tracking process follows a sequential breadth-first approach such that in each step, we predict all the children nodes at the next level of the graph. Each child node is approximately one voxel away from its parent node and is classified into one of three categories: end node, intermediate node, or bifurcation node. The model stops tracking a branch when it predicts an end node. If the node is intermediate, it continues to track the subsequent nodes. For bifurcation nodes, it halts the current branch tracking and assigns a new set of 26 object queries, some of which begin tracking the newly formed branches while the rest are discarded. Trexplorer Super builds on Trexplorer by incorporating the following novel key components. The model architecture is shown in Figure~\ref{fig:arch}.

\subsubsection{Super Trajectory Training. }
Trexplorer generates a centerline graph over 9 steps, starting from the center of a volume patch. New patches are created at the endpoints of the tracked graph, along with the past trajectory, to continue tracking until all endpoints are reached. The past trajectory token, which is an embedding of up to 10 previous node positions, is the only source of past information when tracking in a new patch. This simplifies the formation of training batches as each batch is independent, however, it also results in the loss of the learned past trajectory embedding, stored in object queries, leading to premature branch terminations, lower recall, and missed branches.

To better utilize past trajectory information, we propose Super Trajectory Training (STT). In STT, each sample consists of a super trajectory of size 54, divided into six sub-trajectories of size 10 as they share link nodes. Each sub-trajectory is paired with a volume patch centered at its starting node as shown in Figure \ref{fig:arch}. Secondary branches may appear within a sub-trajectory, but they are only tracked within the corresponding patch during training.

During training, object query outputs from the previous sub-trajectory are used as inputs for the next sub-trajectory, along with new volume patches. This preserves the valuable past trajectory embedded in the object queries, improving tracking. In contrast, Trexplorer's training strategy is analogous to training on a single sub-trajectory. The first sub-trajectory still relies on the past trajectory token. During inference, tracking begins at the root point, and for subsequent patches, the object query outputs from previous patches are used. This forms a continuous chain from the root to all leaf nodes, ensuring more effective tracking.

\subsubsection{Focal Cross Attention. }
In Trexplorer, object queries track branches by aggregating information from image features using the transformer's cross-attention module~\cite{vaswani2017attention}. For tubular structures like vessel trees, capturing long-range dependencies and fine-grained spatial details is crucial for accurately locating thin, elongated branches. However, using large high-resolution features quickly becomes computationally infeasible for 3D medical images. Some methods address this by using learned sparse attention~\cite{zhu2020deformable,roh2021sparse,zheng2023more}, but they rely on object queries to determine which features to attend to. This conflicts with Trexplorer’s use of object queries to store branch tracking history, leading to poor performance.

Trexplorer Super introduces Focal Cross Attention (FCA), which extracts high-resolution features over a large region but restricts cross-attention to the small focal region where branches are being tracked, as shown in Figure \ref{fig:arch}. The responsibility for aggregating long-range dependencies while maintaining fine-grained spatial details is delegated to the feature extractor as the model is trained end-to-end. This design allows the decoder’s object queries to focus on retaining tracking history while cross-attending to a smaller, more relevant set of features.    

\subsubsection{Target Augmentation. }
The main reason for introducing target augmentation is that as the radius of a bifurcation node increases, the area of viable positions for bifurcation also increases, rather than being just a single fixed position. Trexplorer Super is trained to account for this positional ambiguity by using target augmentation. During augmentation, for each bifurcation point in the primary branch of a super trajectory, an offset is sampled from a Laplace distribution with mean $\mu = 0$ and scale $b$ proportional to the bifurcation radius. This offset shifts secondary branches up or down along the primary branch, effectively generating viable augmented targets. To preserve natural trajectories, the branches around these new bifurcation points are smoothed.

Training with augmented targets encourages the model to consider more nodes as potential bifurcations, improving the detection of new branches. This strategy also reduces duplicates, resulting in faster inference and eliminating the need for post-processing. The duplicate reduction can be explained by the fact that combining Super Trajectory Training with Target Augmentation significantly enhances the effect of the Hungarian loss and self-attention between object queries, two key components used for preventing duplicates in Trexplorer. Additionally, a small amount of Gaussian noise $(\mu = 0, \sigma=0.025)$  is added to the rest of the points to further improve robustness.

\section{Experiments and Results}
\subsection{Datasets}
To our knowledge, the only publicly available 3D centerline tree dataset is the synthetic vessel tree dataset~\cite{tetteh2020deepvesselnet}. However, these trees were generated without collision avoidance, leading to self-intersections and intersections between trees, artifacts that do not accurately represent real vessel structures. To address this issue, we use the Synthetic Vascular Toolkit (SVT)~\cite{sexton2023rapid,svt2023} to generate a new synthetic tree dataset with collision avoidance. We follow the same approach as~\cite{tetteh2020deepvesselnet} to create corresponding images. This dataset serves as a useful toy example for model research and development.

For a comprehensive evaluation, we also generate centerline ground truth from two publicly available real tubular tree segmentation datasets: the ATM'22 dataset~\cite{zhang2023multi,zhang2022cfda,zheng2021alleviating,yu2022break,qin2019airwaynet} (airway segmentation) licensed under CC BY-NC and the Parse 2022 dataset~\cite{luo2023efficient} (pulmonary artery segmentation) licensed under CC BY-NC-ND 6.0. The use of these datasets complies with the terms set by the dataset owners. Before extracting ground truth, we resample the data volumes to 0.5mm isotropic resolution. The Vascular Modeling Toolkit (VMTK)~\cite{Izzo2018,slicer3d} is used to extract root points from segmentation masks, which Kimimaro~\cite{silversmith2021kimimaro} then uses to trace the tubular tree centerlines. Further dataset statistics are provided in Table~\ref{tab:data-stats-table}. 

\begin{table}[t!]
\centering
\caption{Statistics for the ground truth centerline graphs of the provided datasets.}
\label{tab:data-stats-table}
{\fontsize{8}{10}\selectfont
\begin{tabular}{c|ccc|c|c|c|c|ccc}
\hline
\multirow{2}{*}{Dataset} & \multicolumn{3}{c|}{Samples} & \multirow{2}{*}{\begin{tabular}[c]{@{}c@{}}Max Node\\ Degree\end{tabular}} & \multirow{2}{*}{\ \begin{tabular}[c]{@{}c@{}}Mean\\ Points\end{tabular}\ } & \multirow{2}{*}{\ \begin{tabular}[c]{@{}c@{}}Mean\\ Depth\end{tabular}\ } & \multirow{2}{*}{\ \begin{tabular}[c]{@{}c@{}}Mean\\ Width\end{tabular}\ } & \multicolumn{3}{c}{Radius} \\ 
                         & train & val. & test & & & & & max  & mean  & min \\ \hline
Synthetic Dataset                 & 368 | & \ 32 | & 100 & 4 & 2763 & 343 & 20 & 17  | & 5.37 | & 2  \\
ATM'22                 & 220 | & \ 16 | & \ 60 & 4 & 7398 & 504 & 55 & 23 | & 2.47 | & 1  \\
Parse 2022                 & \ \  72 | & \ \ 8 | & \ 20 & 4 & 19644 & 498 & 126 & 38 | & 2.55 | & 1  \\
\hline
\end{tabular}}
\end{table} 

\subsection{Evaluation Metrics}
We evaluate predictions at the point, branch, and tree levels to capture different aspects of accuracy. At the point level, we use Precision, Recall, and F1-score. A predicted node is a True Positive (TP) if a ground truth node exists within its 1.5-voxel radius and that has not been matched to another prediction; otherwise, it is a False Positive (FP). An unmatched ground truth node is a False Negative (FN). We also assess radius accuracy using Mean Absolute Error (MAE). Given the large number of nodes, we avoid metrics that require solving the linear assignment problem (LAP).

At the branch level, we use the F1-score but treat branches as objects. A predicted branch is a TP if it correctly matches at least 80\% of the points in a ground truth branch within 1.5 voxels radius, provided the ground truth branch is not already matched. Otherwise, it is an FP. Unmatched ground truth branches are FNs. To evaluate the overall graph structure, we use topological metrics, specifically the MAE of Betti-0 (connected components) and Betti-1 (cycles).

\subsection{Experiments}

We evaluate two previous state-of-the-art (SOTA) centerline tracking and detection models, Vesselformer and Trexplorer, alongside our proposed method, Trexplorer Super, on three datasets. All models were trained on a single node with four A100 GPUs, with Trexplorer and Trexplorer Super using mixed precision for efficiency. To ensure a fair comparison focused on model improvements, Trexplorer and Trexplorer Super share nearly identical hyperparameters: the number of tokens allocated for a bifurcation node is set to 26 and the number of maximum concurrently tracked tokens is set to 196. Both models were trained for approximately 2 million iterations. Vesselformer gave the best results for author-optimized hyperparameters, with 80 object queries per patch and approximately 12 million training iterations. Each model was trained five times per dataset, and we report the mean and standard deviation for each metric.

\subsection{Results}

Table~\ref{tab:res-table-1} and~\ref{tab:res-table-2} presents the results of the three evaluated models on the synthetic and real datasets. On the synthetic dataset, Vesselformer achieves a higher F1-score than Trexplorer, but Trexplorer has better recall, however, it suffers from excessive duplicate branches, leading to low precision. Trexplorer Super improves recall over Vesselformer, though still lower than Trexplorer, while drastically reducing duplicates, resulting in the highest F1-score. It also achieves the best Branch F1-score and lowest radius MAE. Both Trexplorer and Trexplorer Super ensure topological correctness with zero Betti-0 and Betti-1 errors, whereas Vesselformer struggles, predicting multiple disconnected components and cycles.

On ATM'22, Vesselformer retains some centerline tracking ability but misses branches and predicts duplicates, leading to a low overall score. Trexplorer performs even worse, barely tracking any branches due to limited past trajectory information. Trexplorer Super, with its enhancements, achieves a significant performance boost across all metrics. Parse 2022 presents a greater challenge due to denser trees and weaker vessel signals. Both Vesselformer and Trexplorer struggle, but Trexplorer Super outperforms them significantly. However, its performance is significantly lower compared to what it achieves on the other datasets, suggesting that a more powerful pretrained feature extractor could help improve vessel tracking.

Regarding failure cases, we observe that while Trexplorer Super performs well on most test samples, it completely fails on a few. Analyzing the inputs revealed no immediately apparent cause, but one possible factor could be insufficient training data. Further investigation is needed to better understand these failures and improve robustness. 

\begin{figure}[!t]
\centering
\begin{tabular}{>{\centering\arraybackslash}m{.05\linewidth} >{\centering\arraybackslash}m{.225\linewidth} >{\centering\arraybackslash}m{.225\linewidth} >{\centering\arraybackslash}m{.225\linewidth} >{\centering\arraybackslash}m{.225\linewidth}}
    & \textbf{Ground Truth} & \textbf{Vesselformer} & \textbf{Trexplorer} & \textbf{Trexplorer Super} \\

    \rotatebox{90}{\textbf{Synthetic}} & \includegraphics[width=\linewidth,height=\linewidth]{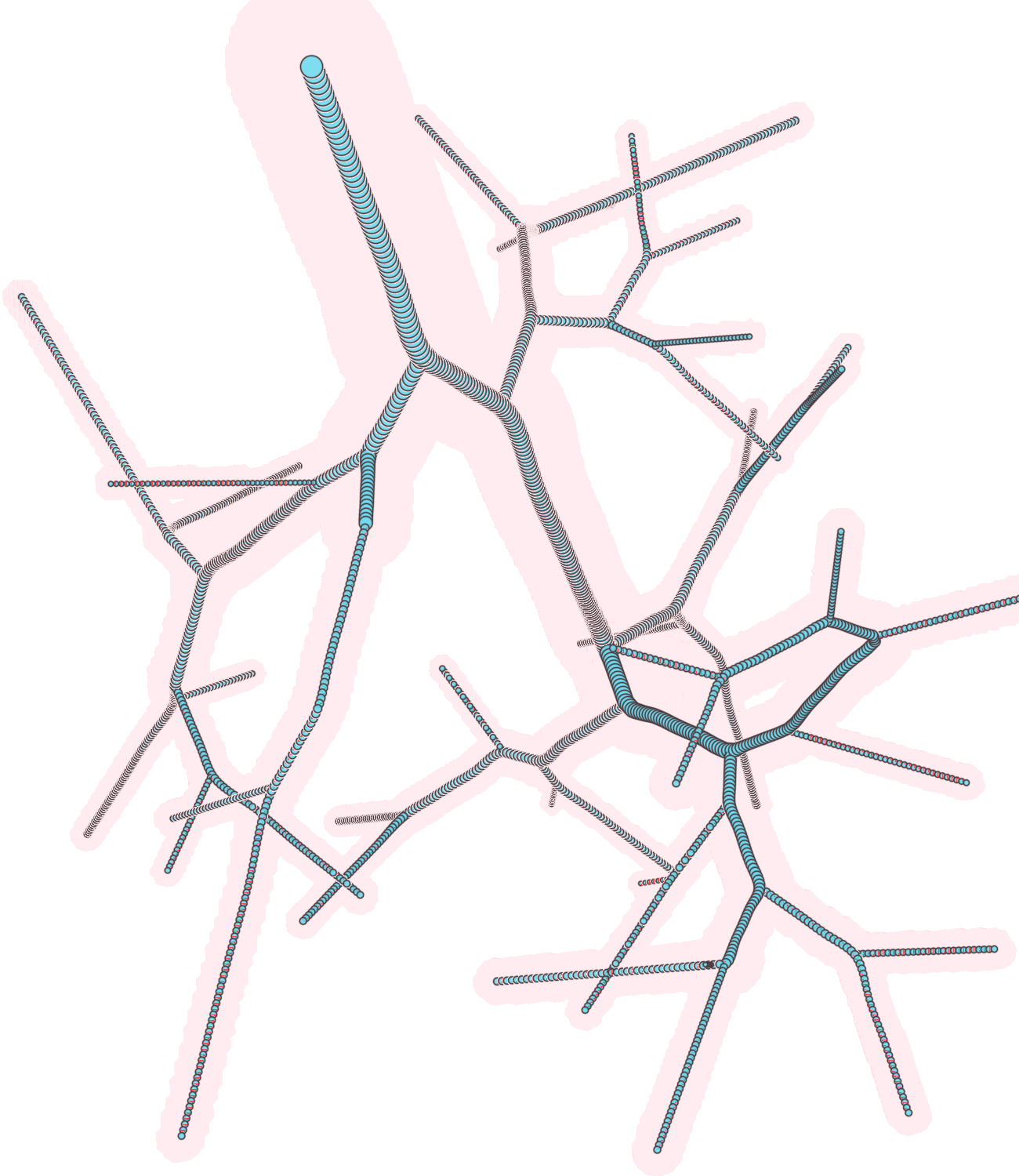} & \includegraphics[width=\linewidth,height=\linewidth]{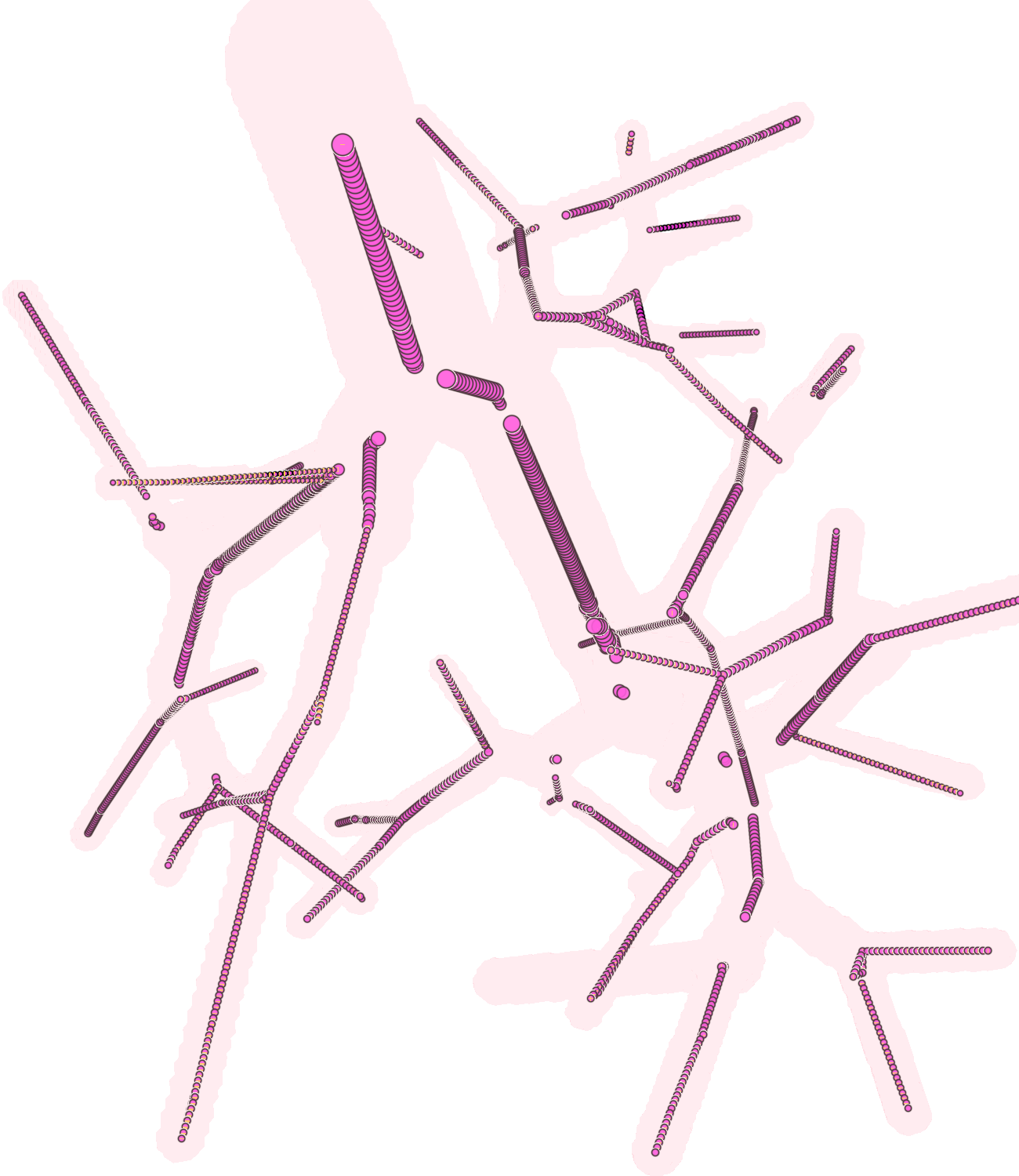} & \includegraphics[width=\linewidth,height=\linewidth]{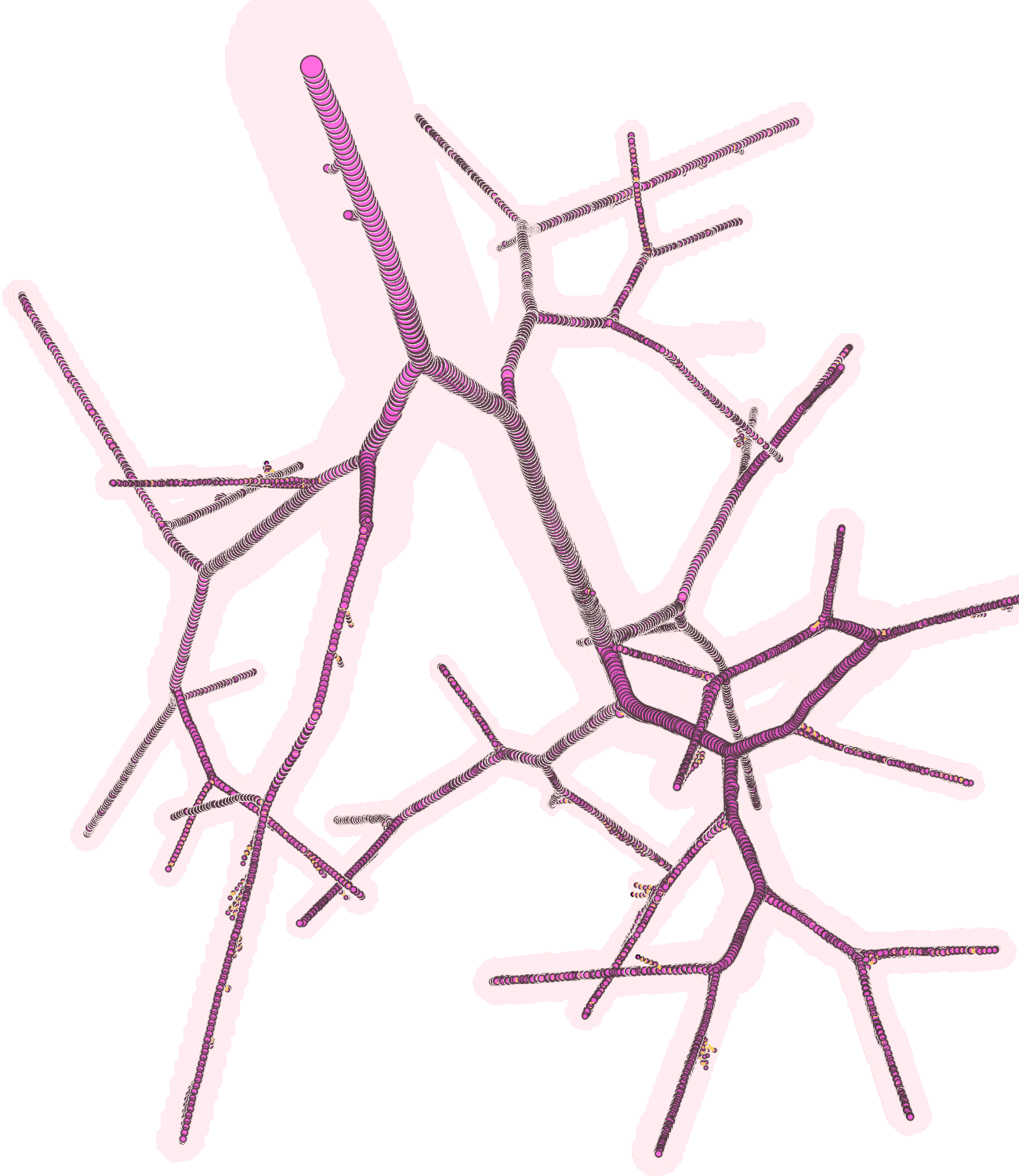} & \includegraphics[width=\linewidth,height=\linewidth]{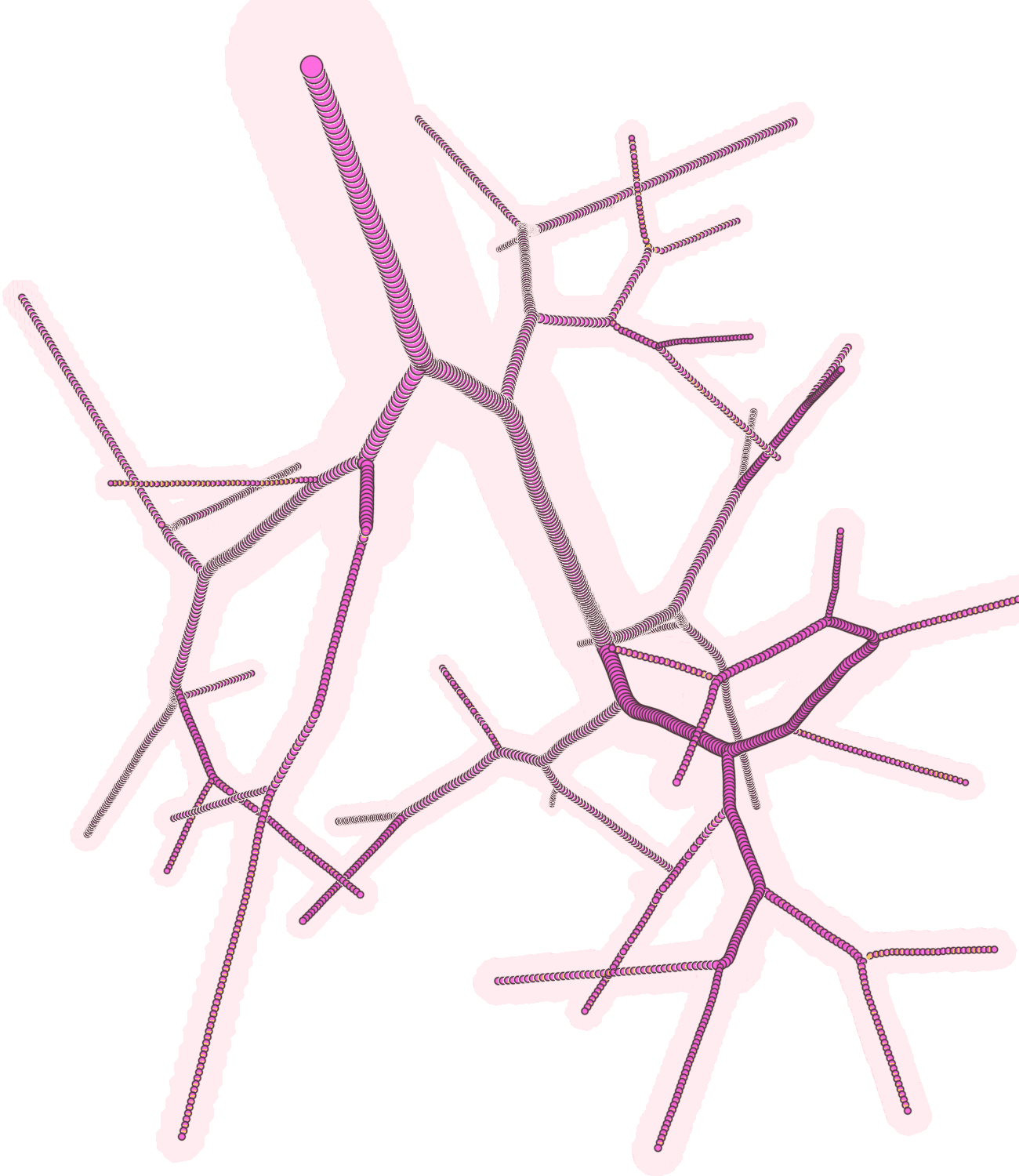} \\

    \rotatebox{90}{\textbf{ATM'22}} & \includegraphics[width=\linewidth,height=\linewidth]{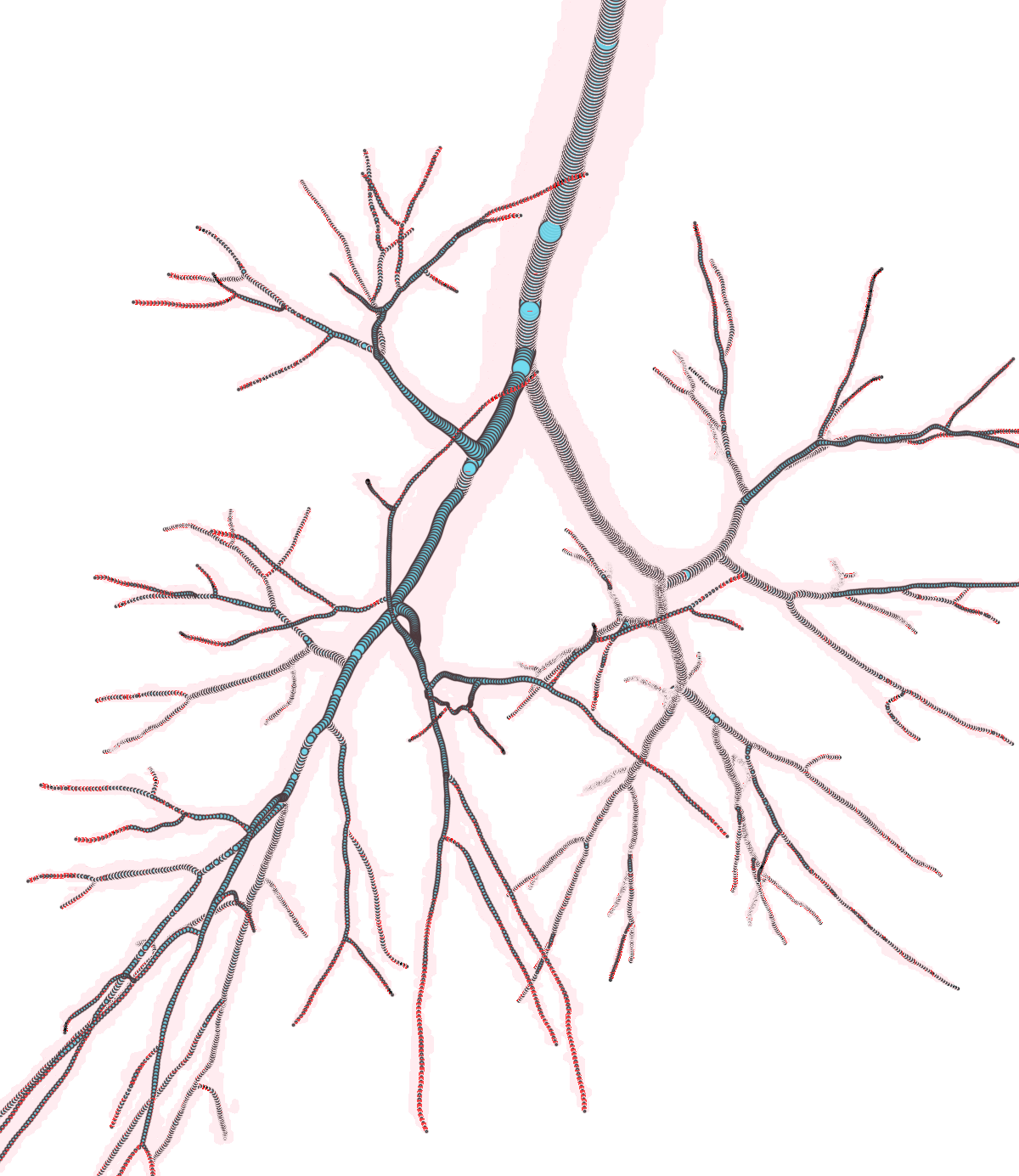} & \includegraphics[width=\linewidth,height=\linewidth]{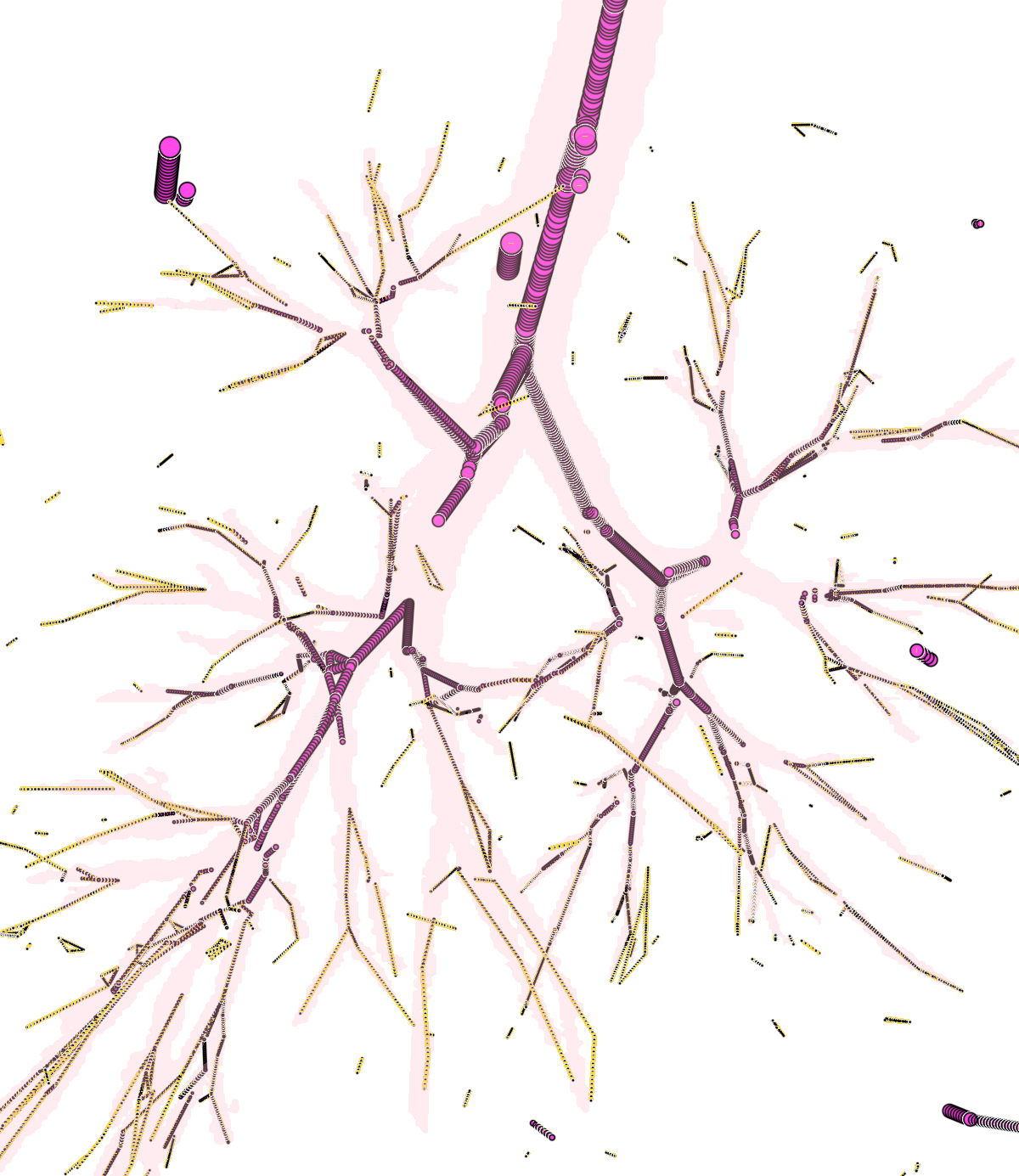} & \includegraphics[width=\linewidth,height=\linewidth]{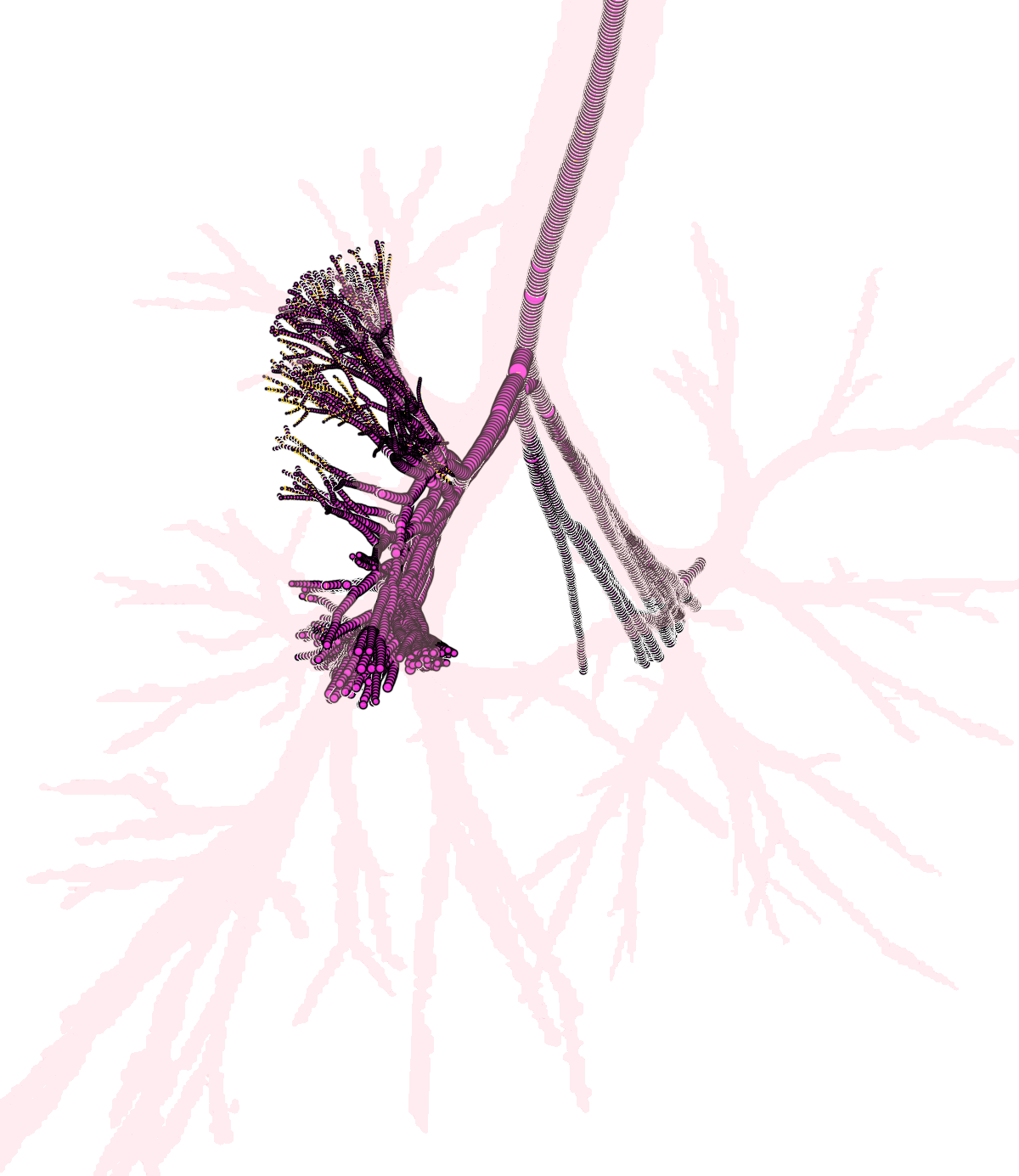} & \includegraphics[width=\linewidth,height=\linewidth]{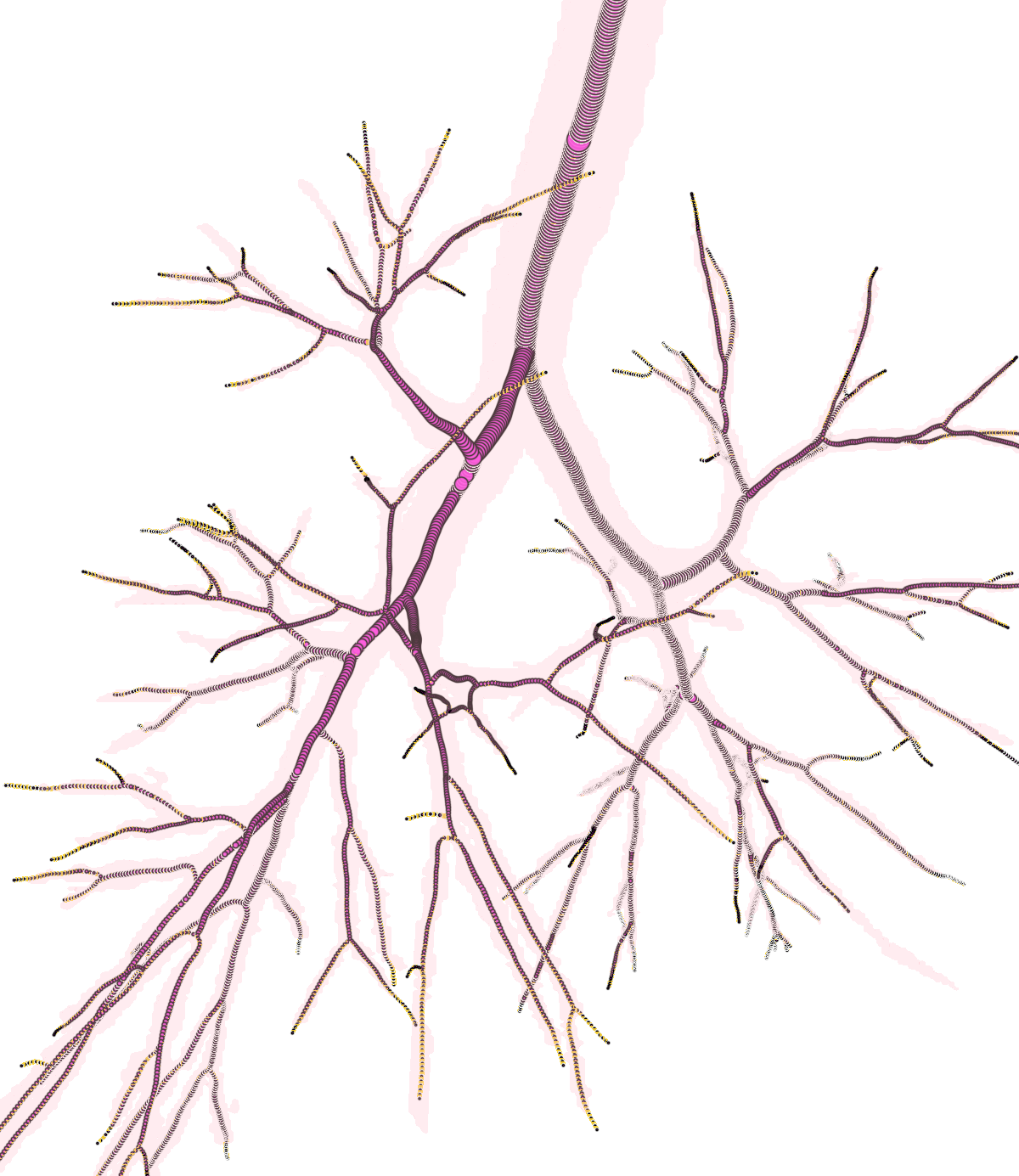} \\

    \rotatebox{90}{\textbf{Parse 2022}} & \includegraphics[width=\linewidth,height=\linewidth]{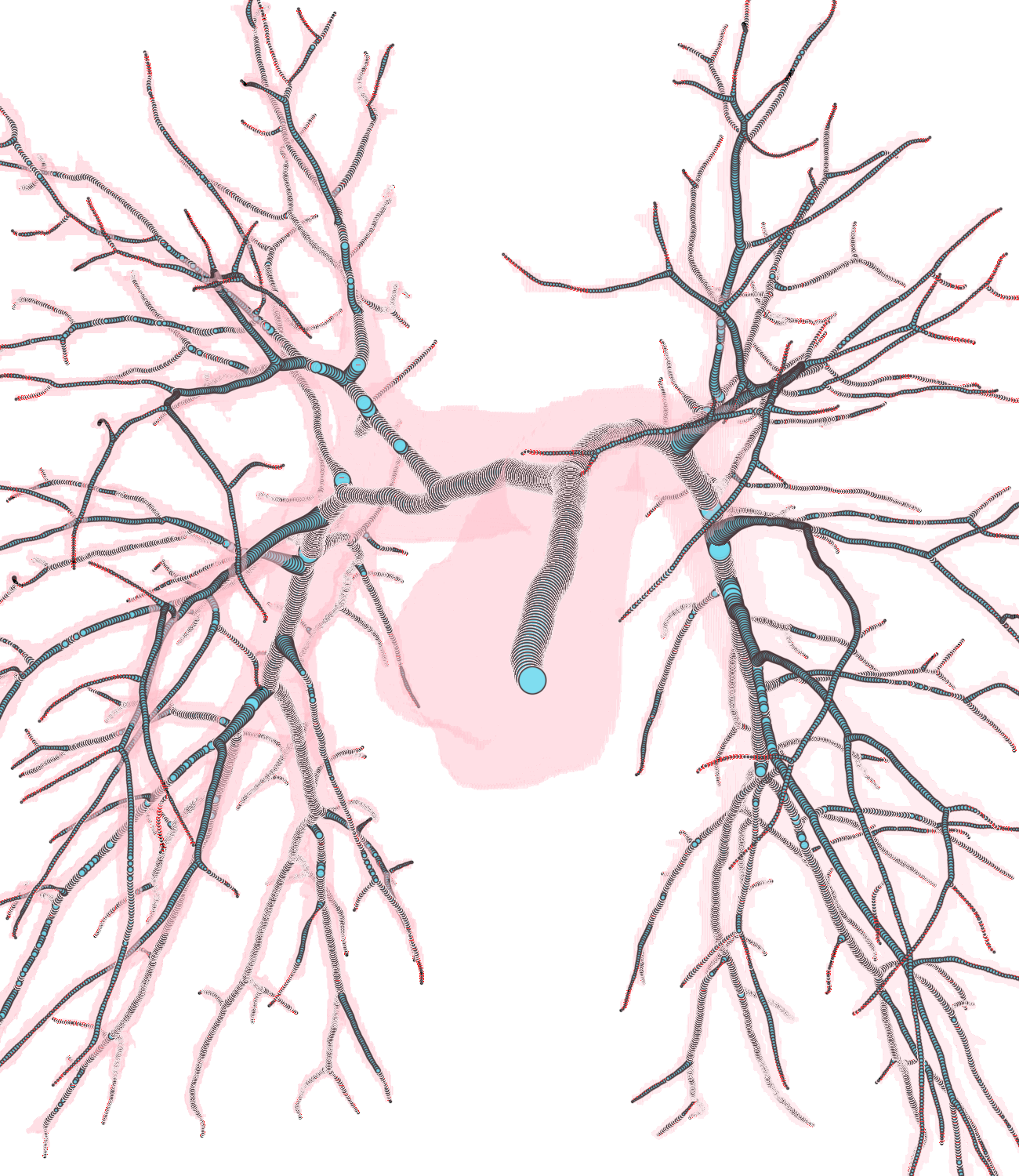} & \includegraphics[width=\linewidth,height=\linewidth]{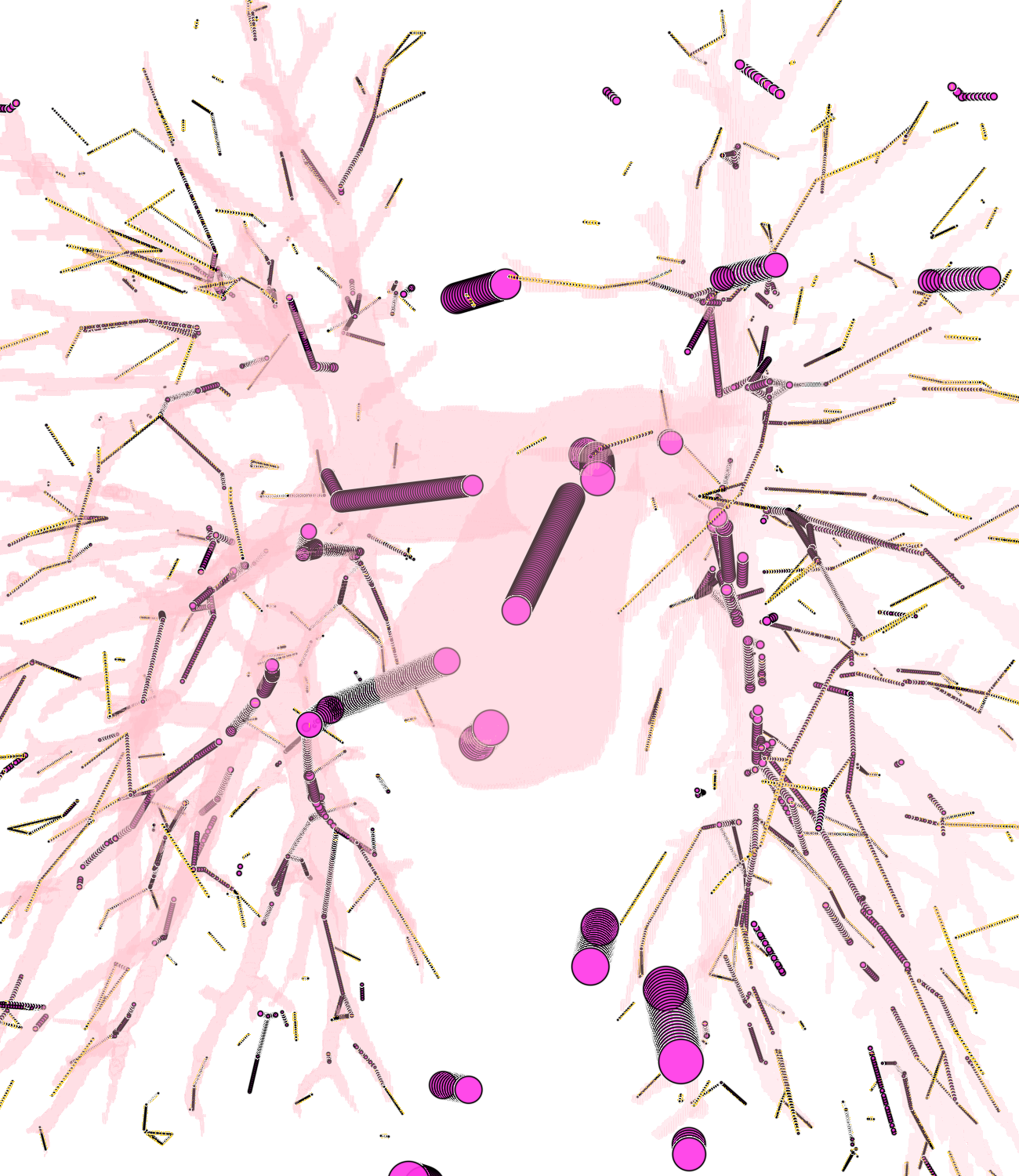} & \includegraphics[width=\linewidth,height=\linewidth]{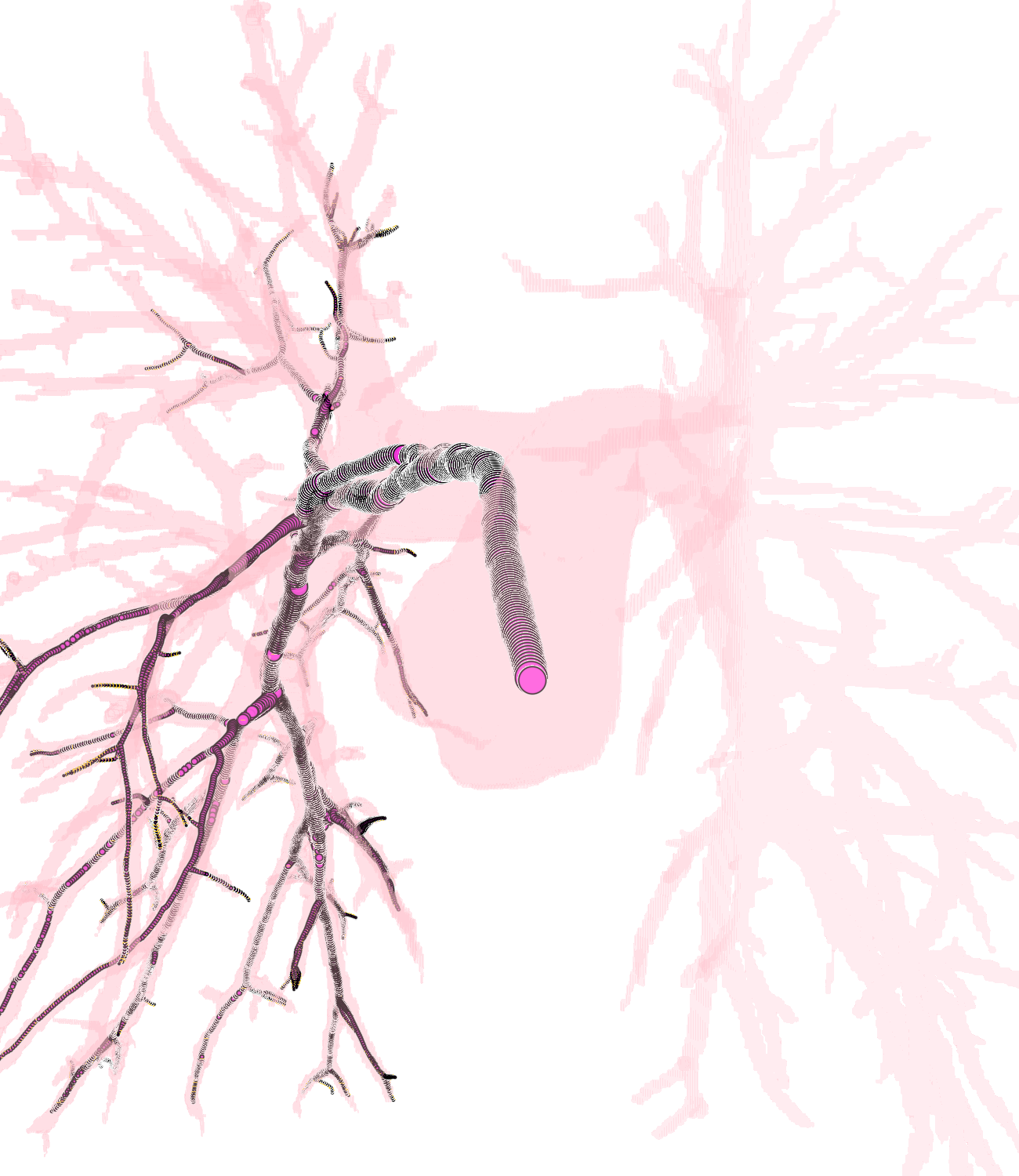} & \includegraphics[width=\linewidth,height=\linewidth]{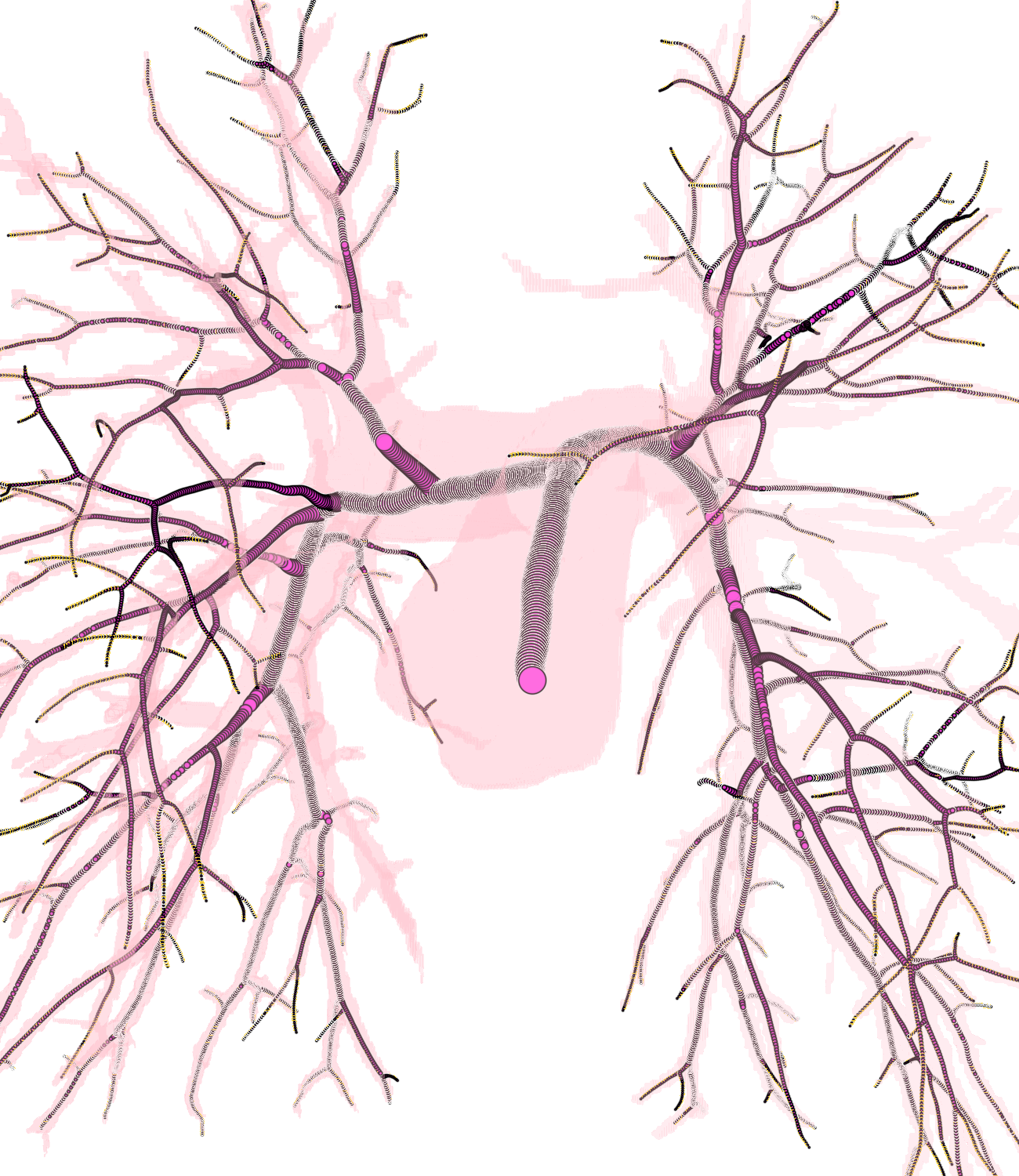} \\
\end{tabular}
\caption{Visual comparison between the ground truth, Vesselformer, Trexplorer, and Trexplorer Super for one sample from each dataset. The centerline marker size is proportional to the radius value at that node.}
\label{fig:comp}
\end{figure}

Figure~\ref{fig:comp} provides a visual comparison of the models on a sample from each dataset.

\begin{table}[!t]
\centering
\caption{Comparison of different models based on Point Metrics for the synthetic, ATM'22, and Parse 2022 datasets.}
\label{tab:res-table-1}
{\fontsize{8}{10}\selectfont
\begin{tabular}{c|cccc}
\hline
\multirow{2}{*}{Model} & \multicolumn{4}{c}{Point Level} \\
                         & Precision(\%)$\uparrow$ & Recall(\%)$\uparrow$ & F1(\%)$\uparrow$ & Radius (MAE)$\downarrow$ \\ \hline
\multicolumn{5}{c}{Synthetic Dataset} \\ \hline
Vesselformer                 & 44.53 ± 7.87  | & 61.52 ± 1.14  | & 48.18 ± 5.62 | & 0.4244 ± 0.0134 \\
Trexplorer                 & 30.91 ± 9.45  | & \textbf{78.21 ± 4.13}  | & 39.40 ± 8.62 | & 0.2263 ± 0.0323 \\
Trexplorer Super                 & \textbf{91.91 ± 3.28}  | & 70.44 ± 3.02  | & \textbf{77.83 ± 1.89} | & \textbf{0.0955 ± 0.0061} \\ \cline{1-5} 
\multicolumn{5}{c}{ATM'22 Dataset} \\ \hline
Vesselformer                 & 22.32 ± 1.35  | & 34.37 ± 0.94  | & 26.77 ± 1.27 | & 0.7908 ± 0.0095 \\
Trexplorer                 & 3.20 ± 0.73  | & 4.33 ± 0.63  | & 3.34 ± 0.30 | & 0.9744 ± 0.0844 \\
Trexplorer Super                 & \textbf{67.51 ± 1.35}  | & \textbf{60.65 ± 2.01}  | & \textbf{60.45 ± 1.03} | & \textbf{0.3925 ± 0.0241} \\ \cline{1-5} 
\multicolumn{5}{c}{Parse 2022 Dataset} \\ \hline
Vesselformer                 & 18.49 ± 1.84  | & 15.28 ± 0.83  | & 16.43 ± 0.78 | & 1.1144 ± 0.0269 \\
Trexplorer                 & 9.87 ± 3.76  | & 12.01 ± 7.46  | & 10.01 ± 4.98 | & 1.2108 ± 0.3042 \\
Trexplorer Super                 & \textbf{55.27 ± 3.00}  | & \textbf{33.99 ± 3.34}  | & \textbf{39.46 ± 1.93} | & \textbf{0.5627 ± 0.0141} \\
\hline
\end{tabular}}
\end{table}

\begin{table}[!t]
\centering
\caption{Comparison of different models based on Branch and Graph Metrics for the synthetic, ATM'22, and Parse 2022 datasets.}
\label{tab:res-table-2}
{\fontsize{8}{10}\selectfont
\begin{tabular}{c|c|cc}
\hline
\multirow{2}{*}{Model} &  \multicolumn{1}{c|}{Branch Level} & \multicolumn{2}{c}{Graph Level (MAE)} \\ 
                         & F1(\%)$\uparrow$  & Betti-0$\downarrow$ & Betti-1$\downarrow$ \\ \hline
\multicolumn{4}{c}{Synthetic Dataset} \\ \hline
Vesselformer                 & 15.95 ± 0.36 & 81.7 ± 16.8   | & 653.5 ± 138.7  \\
Trexplorer                 & 26.26 ± 7.18 & \textbf{0.000 ± 0.0}   | & \textbf{0.000 ± 0.0}  \\
Trexplorer Super                 & \textbf{77.12 ± 1.59}  & \textbf{0.000 ± 0.0}   | & \textbf{0.000 ± 0.0}  \\ \cline{1-4} 
\multicolumn{4}{c}{ATM'22 Dataset} \\ \hline
Vesselformer                 & 1.95 ± 0.25 & 312.5 ± 25.1   | & 180.4 ± 35.9  \\
Trexplorer                 & 3.34 ± 0.30 & \textbf{0.00 ± 0.0}   | & \textbf{0.00 ± 0.0}  \\
Trexplorer Super                 & \textbf{41.15 ± 1.28}  & \textbf{0.00 ± 0.0}   | & \textbf{0.00 ± 0.0}  \\ \cline{1-4} 
\multicolumn{4}{c}{Parse 2022 Dataset} \\ \hline
Vesselformer                 & 1.99 ± 0.16 & 410.1 ± 23.9   | & 246.7 ± 78.1 \\
Trexplorer                 & 3.71 ± 1.91 & \textbf{0.00 ± 0.0}   | & \textbf{0.00 ± 0.0}  \\
Trexplorer Super                 & \textbf{23.46 ± 1.09}  & \textbf{0.00 ± 0.0}   | & \textbf{0.00 ± 0.0}  \\
\hline
\end{tabular}}
\end{table}

\subsection{Ablations}
We conduct an ablation study on the ATM’22 dataset to evaluate the impact of our key modifications to Trexplorer: Super Trajectory Training, Focal Cross Attention, and Target Augmentation. Table \ref{tab:ablation-table} reports the mean and standard deviation of point metrics: precision, recall, and F1-score averaged over three runs per ablation. Our results highlight Super Trajectory Training as the most crucial improvement, allowing Trexplorer to perform effectively on real data. Focal cross-attention further enhances performance by enabling the feature extractor to condense relevant information in the focal region. Target augmentation improves both recall and precision by reducing duplicate predictions and increasing new branch detections, leading to more complete branch reconstructions.

\begin{table}[!t]
\centering
\caption{Evaluation of the novel key components, Super Trajectory Training (STT), Focal Cross Attention (FCA), and Target Augmentation (TA) incorporated in the Trexplorer framework by Trexplorer Super for the ATM'22 Dataset.}
\label{tab:ablation-table}
{\fontsize{8}{10}\selectfont
\begin{tabular}{c|ccc|ccc}
\hline
\multirow{2}{*}{Index} & \multicolumn{3}{c|}{Novel Key Components} & \multicolumn{3}{c}{Point Metrics} \\ 
                         &\ \ \  STT \ \ \ \  & FCA \ \ \ \  & TA \ \ \ & Precision$\uparrow$  & Recall$\uparrow$ & F1$\uparrow$ \\ \hline
1                 &  &  &  & 0.0307 ± 0.0008 | & 0.0359 ± 0.0021 | & 0.0317 ± 0.0011 \\
2                 &  & \checkmark &  & 0.3657 ± 0.5494 | & 0.0366 ± 0.0320 | & 0.0256 ± 0.0219 \\
3                 & \checkmark &  &  & 0.4487 ± 0.1020 | & 0.2743 ± 0.0769 | & 0.3301 ± 0.0870 \\
4                 & \checkmark & \checkmark &  & 0.6176 ± 0.0223 | & 0.5472 ± 0.0745 | & 0.5391 ± 0.0505 \\
5                 & \checkmark & \checkmark & \checkmark & \textbf{0.6782 ± 0.0104} | & \textbf{0.6120 ± 0.0250} | & \textbf{0.6066 ± 0.0064}\\

\hline
\end{tabular}}
\end{table}

\section{Conclusion}
We present major improvements to Trexplorer, enhancing its accuracy, robustness, and completeness. Additionally, we introduce three datasets and conduct a comprehensive evaluation, showing our model outperforms two state-of-the-art baselines. While results are promising, challenges remain, particularly on the Parse 2022 dataset. Future work includes leveraging a larger pretrained feature extractor and integrating advanced DETR variants or recurrent architectures like LSTMs for further refinement.

\subsubsection{\discintname}
The authors have no competing interests to declare that are relevant to the content of this article.

\bibliographystyle{splncs04}
\bibliography{trexplorer_super}

\end{document}